\documentclass{article}
\usepackage{spconf,amsmath,graphicx,hyperref}
\usepackage{amssymb,booktabs,multirow,makecell,enumitem}
\usepackage[table]{xcolor} 

\title{Dynamic Spectrum Re-parameterized Convolution for Cost-free Image Enhancement}

\name{Zhiyan Zhou$^{1}$, Hailong Yan$^{1}$, Hanlin Liu, Yongrui Zhang, Xiangtao Zhang, Le Zhang
\thanks{$^{1}$Equal Contribution.} 
\address{University of Electronic Science and Technology of China}}

\let\oldthebibliography\thebibliography
\let\endoldthebibliography\endthebibliography

\renewenvironment{thebibliography}[1]
  {\oldthebibliography{#1}%
   \setlength{\itemsep}{0pt}%
   \setlength{\parsep}{0pt}}
  {\endoldthebibliography}
  
\begin{document}
%
\maketitle
\begin{abstract}
Image enhancement spans degradations with distinct spectral characteristics. Standard convolutions use fixed kernels, whereas dynamic convolutions gain adaptability at additional inference cost. We propose Dynamic Spectrum Re-parameterized Convolution (DSRC), a plug-and-play module that learns task-adaptive spectral responses during training while retaining standard convolution at inference. DSRC decomposes convolution kernels into frequency bands, applies learnable high/low-pass modulation, and incorporates lightweight residual branches. Owing to the linearity of these operations, all training-time components are analytically fused into a single kernel for inference. Experiments on low-light enhancement, underwater enhancement, and image super-resolution show consistent PSNR gains and generally improved or preserved SSIM and LPIPS across twelve backbone–task configurations, with unchanged deployed parameter counts and theoretical FLOPs, and no additional inference operators.
\end{abstract}

\begin{keywords}
Image enhancement, Cost-free operation, Convolutional neural network
\end{keywords}

\section{Introduction}
\label{sec:intro}

Image Enhancement (IE) tasks such as Low-Light Image Enhancement (LLIE), Underwater Image Enhancement (UIE), and image Super-Resolution (SR) aim to restore high-quality images under degradations, thereby enhancing both perceptual quality and the efficacy of downstream vision tasks.

    \begin{figure}
        \centering
        \includegraphics[width=1\linewidth]{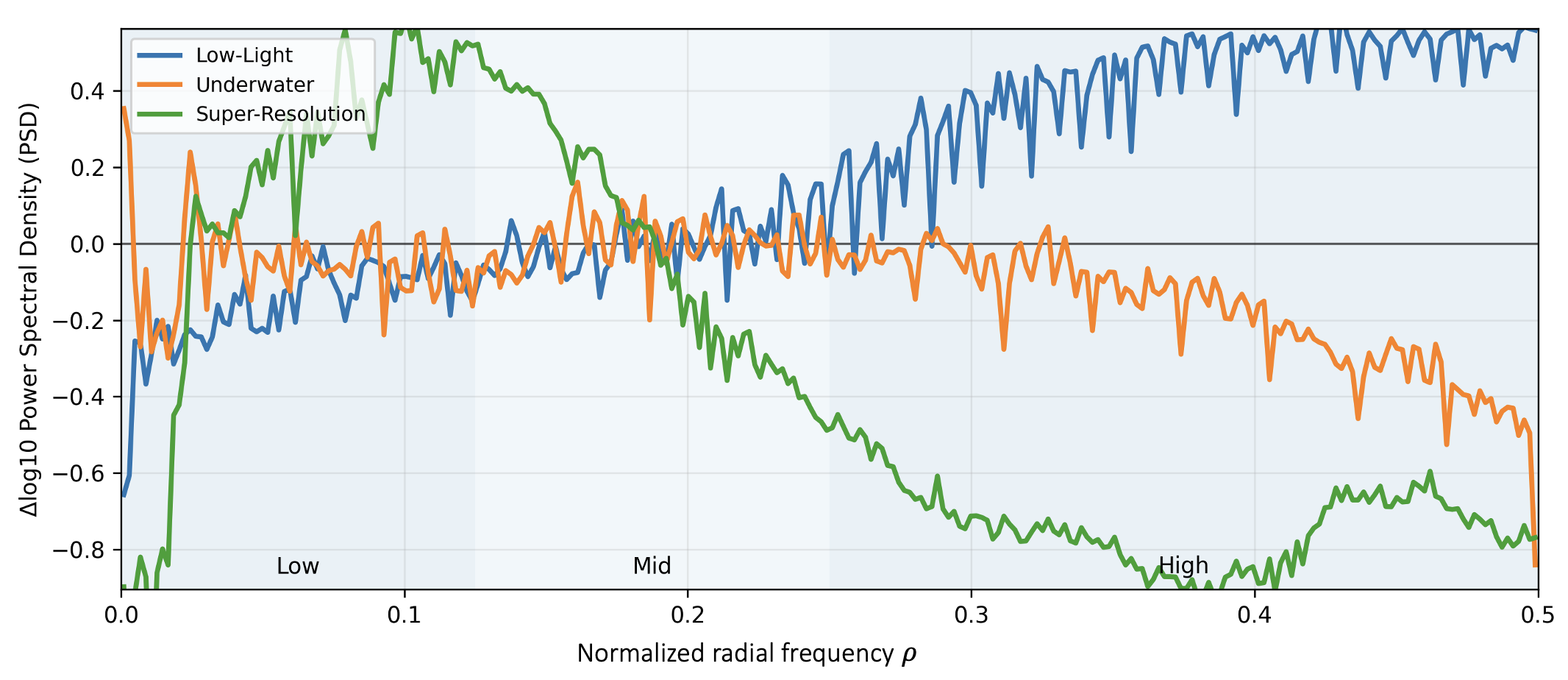}
        \caption{Relative spectral profiles across image enhancement tasks. Each curve shows the input-minus-reference difference in $\Delta\log_{10}\mathrm{PSD}$, centered by its mean over [1/8, 1/4). Positive and negative values indicate larger and smaller spectral ratios relative to this middle-band baseline.
}
        \label{fig:freq}
    \end{figure}

    \begin{figure*}[t]
    \centering
    \includegraphics[width=\linewidth]{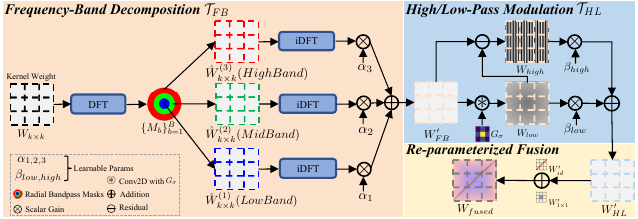}
    \caption{Overview of DSRC. Kernel weights are decomposed into fixed frequency bands with learnable gains, modulated by high/low-pass components, and fused with residual branches into a standard $W_{fused}$ for zero-overhead inference.}
    \label{fig:main}
    \end{figure*}

Convolutional Neural Network (CNN)-based methods \cite{qin2020ffa,wu2021contrastive,chen2018learning,yan2025igdnet} offer a favorable accuracy–efficiency trade-off. Conventional convolutions learn their weights directly in the spatial domain, without explicitly parameterizing the contributions of different frequency bands. Transformer-based \cite{choi2024reciprocal,li2023feature} and diffusion-based methods \cite{xia2023diffir,yi2024diffraw} improve restoration quality through long-range modeling or generative priors, but their attention computation or iterative sampling can be expensive on resource-constrained devices. Dynamic convolutions \cite{chen2020dynamic,yang2019condconv} directly adapt kernel responses to input content, yet this flexibility is typically accompanied by additional inference parameters and computation.

Structural re-parameterization \cite{cai2025refconv} offers a complementary route: multi-branch structures used during training are analytically merged into a single kernel for deployment. Existing re-parameterized convolutions, however, are largely agnostic to the degradation-dependent spectral behavior of enhancement tasks. Frequency-aware methods model spectral information through feature fusion \cite{chen2024frequency}, spectral training objectives \cite{wei2025freqinr}, channel-specific enhancement \cite{yan2026revisiting}, and frequency-decoupled feature manipulation in image generation \cite{yan2026morphsig}. DSRC instead directly parameterizes convolution kernels for analytic fusion.

Fig.~\ref{fig:freq} compares the centered spectral differences between input and reference images across three enhancement tasks. In the displayed profiles, LLIE shows predominantly positive high-frequency differences, whereas UIE and SR show predominantly negative differences over the upper frequency range. These signs describe spectral ratios relative to the middle-band baseline, rather than absolute energy excess or deficiency. The observed variation motivates a spectrum-aware kernel parameterization that learns spectral weightings during training.

We propose Dynamic Spectrum Re-parameterized Convolution (DSRC), which learns kernel-level spectral responses during training and folds all components into a standard inference kernel. DSRC first decomposes a kernel into fixed radial frequency bands with learnable gains, then modulates its low- and high-frequency components, and finally adds lightweight residual compensation. The spectral parameters are learned during training and absorbed into the deployed kernel; they are shared across inputs rather than predicted for each image. Our contributions are:
\begin{itemize}[leftmargin=*,nosep]
    \item We introduce a plug-and-play spectral convolution unit whose training-time branches collapse into a standard convolution with zero additional inference overhead.
    \item We combine FB decomposition, HL modulation, and residual compensation (RC) to accommodate degradation-specific spectral requirements.
    \item We evaluate DSRC on three IE tasks and twelve backbone–task configurations, where it consistently improves PSNR while generally preserving or improving perceptual and structural metrics.
\end{itemize}

\section{Proposed Method}
\label{sec:METHOD}

    \begin{table*}[t]
    \centering
    \caption{Performance comparison on three image enhancement tasks. Methods marked with * replace standard convolutions with DSRC. ↑: The higher, the better, ↓: the lower, the better. Params(K), FLOPs(G), and Latency(ms).}
    \label{tab:comparison}
    \resizebox{\textwidth}{!}{
    \begin{tabular}{cccc|ccc|cccc|ccc|cccc|ccc}
    \toprule
    \multicolumn{7}{c|}{\textbf{Low-Light Image Enhancement}} & \multicolumn{7}{c|}{\textbf{Underwater Image Enhancement}} & \multicolumn{7}{c}{\textbf{Image Super-Resolution}}\\
    \cmidrule(lr){1-7}\cmidrule(lr){8-14}\cmidrule(lr){15-21}
    \textbf{Methods}
    & PSNR $\uparrow$ & SSIM $\uparrow$ & LPIPS $\downarrow$ & Params $\downarrow$ & FLOPs $\downarrow$ & Latency $\downarrow$  
    & \textbf{Methods}
    & PSNR $\uparrow$ & SSIM $\uparrow$ & LPIPS $\downarrow$ & Params $\downarrow$ & FLOPs $\downarrow$ & Latency $\downarrow$ 
    & \textbf{Methods}
    & PSNR $\uparrow$ & SSIM $\uparrow$ & LPIPS $\downarrow$ & Params $\downarrow$ & FLOPs $\downarrow$ & Latency $\downarrow$  \\
    \midrule
    MobileIE \cite{yan2025mobileie}   & 23.01 & 0.80 & 0.30 & 4.05 & 0.92 & 1.82 &
    MobileIE \cite{yan2025mobileie}   & 23.63 & 0.90 & 0.18 & 4.05 & 1.18 & 2.06 &
    IDN \cite{hui2018fast}            & 37.57 & 0.95 & 0.09 & 552.76 & 39.74 & 20.62 \\
    \rowcolor{gray!15}
    \textbf{MobileIE*}  & \textbf{23.74} & \textbf{0.82} & \textbf{0.29} & \textbf{(+0)} & \textbf{(+0)} & \textbf{(+0)} &
    \textbf{MobileIE*}  & \textbf{23.86} & \textbf{0.91} & \textbf{0.17} & \textbf{(+0)} & \textbf{(+0)} & \textbf{(+0)} &
    \textbf{IDN*}       & \textbf{37.73} & \textbf{0.96} & \textbf{0.09} & \textbf{(+0)} & \textbf{(+0)} & \textbf{(+0)} \\
    \midrule
    LYT-Net \cite{brateanu2025lyt}    & 25.82 & 0.83 & 0.14 & 44.92 & 3.65 & 6.31 &
    GHSUIR \cite{guo2025underwater}   & 23.29 & 0.91 & 0.11 & 1144.82 & 20.10 & 92.52 &
    LAPSRN \cite{lai2017deep}         & 37.43 & 0.95 & 0.09 & 288.34 & 39.11 & 9.07 \\
    \rowcolor{gray!15}
    \textbf{LYT-Net*}   & \textbf{26.05} & \textbf{0.84} & \textbf{0.14} & \textbf{(+0)} & \textbf{(+0)} & \textbf{(+0)} &
    \textbf{GHSUIR*}   & \textbf{23.57} & \textbf{0.92} & \textbf{0.10} & \textbf{(+0)} & \textbf{(+0)} & \textbf{(+0)} &
    \textbf{LAPSRN*}    & \textbf{37.57} & \textbf{0.96} & \textbf{0.09} & \textbf{(+0)} & \textbf{(+0)} & \textbf{(+0)} \\
    \midrule
    SYENet \cite{gou2023syenet}       & 21.24 & 0.79 & 0.30 & 102.47 & 22.14 & 21.12 &
    UIEC2Net \cite{wang2021uiec}      & 23.89 & 0.91 & 0.12 & 534.96 & 51.95 & 321.98 &
    ESPCN \cite{shi2016real}          & 36.45 & 0.95 & 0.10 & 21.28 & 1.38 & 5.10 \\
    \rowcolor{gray!15}
    \textbf{SYENet*}    & \textbf{22.13} & \textbf{0.81} & \textbf{0.28} & \textbf{(+0)} & \textbf{(+0)} & \textbf{(+0)} &
    \textbf{UIEC2Net*}  & \textbf{24.56} & \textbf{0.92} & \textbf{0.10} & \textbf{(+0)} & \textbf{(+0)} & \textbf{(+0)} &
    \textbf{ESPCN*}     & \textbf{36.62} & \textbf{0.95} & \textbf{0.09} & \textbf{(+0)} & \textbf{(+0)} & \textbf{(+0)} \\
    \midrule
    Zero-DCE \cite{guo2020zero}       & 17.57 & 0.55 & 0.41 & 79.41 & 19.01 & 7.13 &
    UWCNN \cite{li2020underwater}     & 19.38 & 0.86 & 0.20 & 39.97 & 5.22 & 1.72 &
    SRCNN \cite{dong2015image}        & 36.67 & 0.95 & 0.10 & 24.51 & 6.40 & 5.94 \\
    \rowcolor{gray!15}
    \textbf{Zero-DCE*}  & \textbf{17.80} & \textbf{0.56} & \textbf{0.40} & \textbf{(+0)} & \textbf{(+0)} & \textbf{(+0)} &
    \textbf{UWCNN*}     & \textbf{19.62} & \textbf{0.86} & \textbf{0.19} & \textbf{(+0)} & \textbf{(+0)} & \textbf{(+0)} &
    \textbf{SRCNN*}     & \textbf{36.70} & \textbf{0.96} & \textbf{0.10} & \textbf{(+0)} & \textbf{(+0)} & \textbf{(+0)} \\
    \bottomrule
    \end{tabular}}
    \end{table*}

Given a degraded image $x$, we predict $\hat y=f_{\theta}(x)$ by replacing compatible odd-sized backbone convolutions with DSRC. As shown in Fig.~\ref{fig:main}, DSRC combines frequency-band (FB) decomposition, high/low-pass (HL) modulation, and residual compensation (RC), which are fused before inference.

\subsection{Spectral Profile Learning via Frequency-Band Decomposition}
A 2D Discrete Fourier Transform (DFT) is first applied to the spatial kernel $W_{k\times k}\in\mathbb{R}^{C_{out}\times C_{in}\times k\times k}$:
\begin{equation}
    \hat W=\mathcal{F}(W_{k\times k}).
\end{equation}
Let $f_x,f_y\in[-0.5,0.5)$ be normalized frequency coordinates. We define the normalized Euclidean radial frequency as
\begin{equation}
    \rho(f_x,f_y)=\frac{\sqrt{f_x^2+f_y^2}}{\sqrt{2}}.
\end{equation}
For monotonically increasing boundaries $\{e_b\}_{b=0}^{B}$ with $e_0=0$ and $e_B=0.5$, mutually exclusive masks are
\begin{equation}
    M_b(f_x,f_y)=\mathbf{1}[e_{b-1}\leq\rho(f_x,f_y)<e_b].
\end{equation}
The masked sub-band kernels and their weighted reconstruction are
\begin{equation}
    \hat W_b=\hat W\odot M_b,\quad W_b=\operatorname{real}(\mathcal{F}^{-1}(\hat W_b)).
\end{equation}
\begin{equation}
    W'_{FB}=\mathcal{T}_{FB}(W_{k\times k})=\sum_{b=1}^{B}\alpha_b W_b.
\end{equation}
where $\alpha_b$ are learnable scalar gains that adapt the kernel response to different spectral regions. The radial construction makes each mask symmetric with respect to positive and negative frequencies; therefore, for a real spatial kernel, masking preserves conjugate symmetry and the inverse transform is real up to numerical precision. Unlike feature-level frequency modules, this operation reshapes the convolution kernel itself and can consequently be absorbed into the final re-parameterized weight. For an odd kernel size $k$, the discrete DFT grid satisfies $\rho\leq(k-1)/(2k)<0.5$. The masks therefore cover all sampled frequencies without overlap, giving $\sum_b M_b=1$. Hence, setting all $\alpha_b=1$ reconstructs the original kernel, making the unmodulated convolution a special case of FB decomposition. Learning $\alpha_b$ adjusts the contribution of each spectral region without requiring a separate feature transform.

    \begin{figure}
        \centering
        \includegraphics[width=1\linewidth]{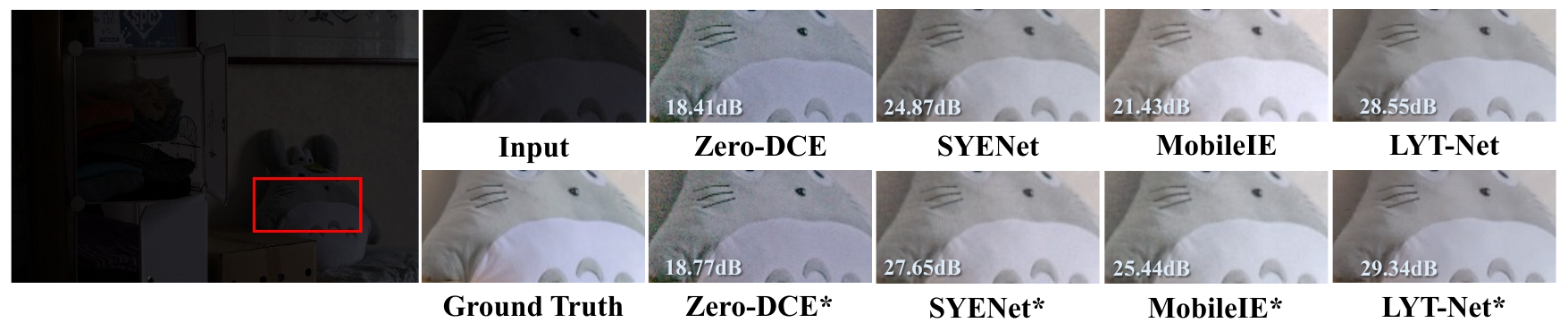}
        \includegraphics[width=1\linewidth]{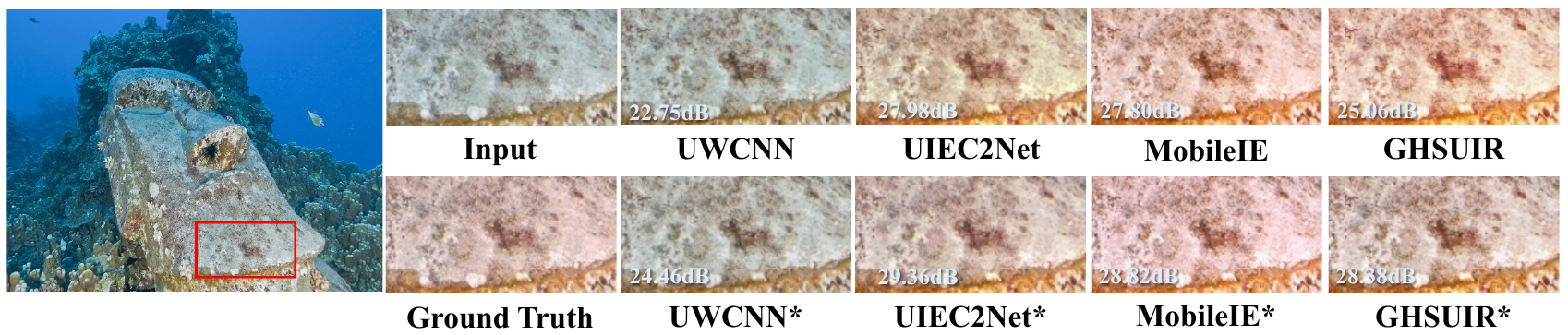}
        \includegraphics[width=1\linewidth]{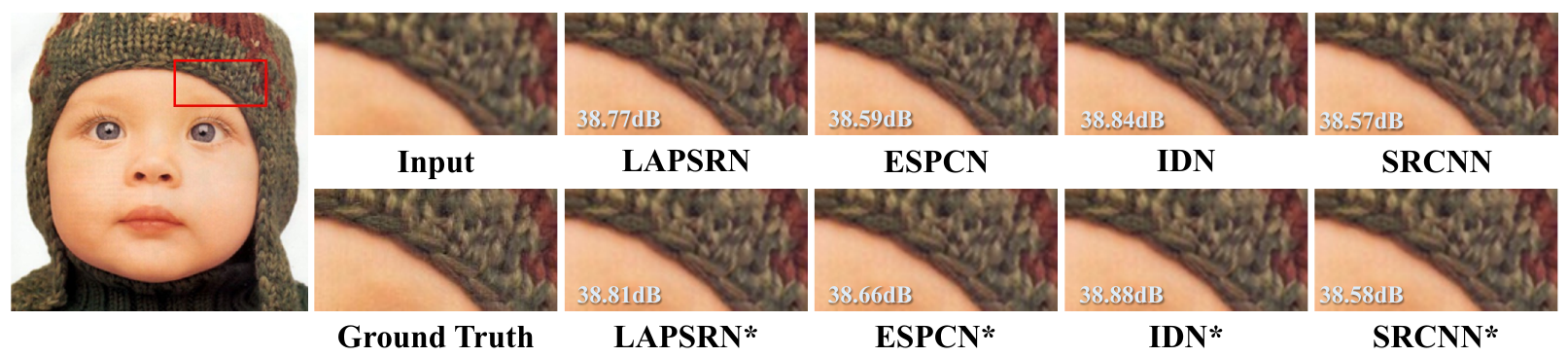}
        \caption{Visual comparison for LLIE (top), UIE (middle), and SR (bottom). PSNR is annotated above each result. DSRC variants are consistently closer to the ground truth than their corresponding baselines.}
        \label{fig:qualitative}
    \end{figure}

    \subsection{High/Low-Pass Modulation}
For an odd kernel size $k$, we define a centered isotropic Gaussian kernel
\begin{equation}
G_{\sigma}(i,j)=\frac{1}{Z}\exp\!\left(-\frac{i^2+j^2}{2\sigma^2}\right),\qquad \sum_{i,j}G_\sigma(i,j)=1.
\end{equation}
Applying $G_\sigma$ independently to each spatial kernel with same padding gives
\begin{equation}
W_{low}=W'_{FB}*G_\sigma,\qquad W_{high}=W'_{FB}-W_{low}.
\end{equation}
and the modulated kernel is
\begin{equation}
W'_{HL}=\mathcal{T}_{HL}(W'_{FB})=\beta_{low}W_{low}+\beta_{high}W_{high}.
\end{equation}
where $\beta_{low}$ and $\beta_{high}$ are learnable gains for the smoothed kernel and its residual, respectively. Gaussian smoothing reduces local variation in the kernel coefficients, while $W_{high}$ retains the component removed by smoothing. This decomposition allows training to adjust the two contributions without assigning them fixed roles such as noise suppression or detail recovery. Since $W_{low}+W_{high}=W'_{FB}$, setting both gains to one recovers the FB kernel exactly. HL therefore includes FB as a special case.
  
    \subsection{Re-parameterized Fusion}

DSRC branches can be fused once training is complete, provided their outputs share the same spatial grid and compatible channel connectivity, with no intervening nonlinearities. The identity branch requires preserved spatial dimensions, unit stride, and matching input and output channels. Branch biases, if present, are summed. The $1\times1$ residual filter is embedded at the center of a $k\times k$ kernel,
\begin{equation}
W'_{1\times1}=W_{1\times1}\cdot\mathbf{I}_{center}.
\end{equation}
while, for shape-preserving convolutions, the identity branch is represented as
\begin{equation}
W'_{id}=\lambda\mathbf{I}_{identity}.
\end{equation}
where $I_{\rm identity}$ denotes the centered channel-wise identity kernel, $\lambda$ is learnable, and $W'_{id}=0$ when the identity branch is absent. The equivalent inference kernel is
\begin{equation}\label{eq:total}
W_{fused}=W'_{HL}+W'_{1\times1}+W'_{id}.
\end{equation}
The $1\times1$ branch supplies localized channel mixing at the kernel center, while the identity branch preserves a direct channel-wise path. These branches are used only in the training-time parameterization. Thus, spectral modulation and residual compensation are reduced exactly to a standard convolution at inference. The deployed network therefore keeps the parameter count, FLOPs, and operator structure of the corresponding baseline convolution. Importantly, ``zero-overhead'' refers to inference rather than training: DSRC introduces learnable band gains, modulation gains, and residual branches during optimization, but none of these remain as separate operators after re-parameterization. The DFT-based decomposition and Gaussian filtering are applied to kernel weights during training/fusion, not to input feature maps at deployment. Consequently, inference requires neither FFT operations nor input-dependent branch selection. Once $W_{fused}$ is constructed, the forward pass is identical in form to the original convolution, which explains the unchanged inference complexity reported in Tables~\ref{tab:comparison} and~\ref{tab:similar}.

\section{Experiments}
\label{sec:experiments}

    \begin{figure}
    \centering
    \includegraphics[width=1\linewidth]{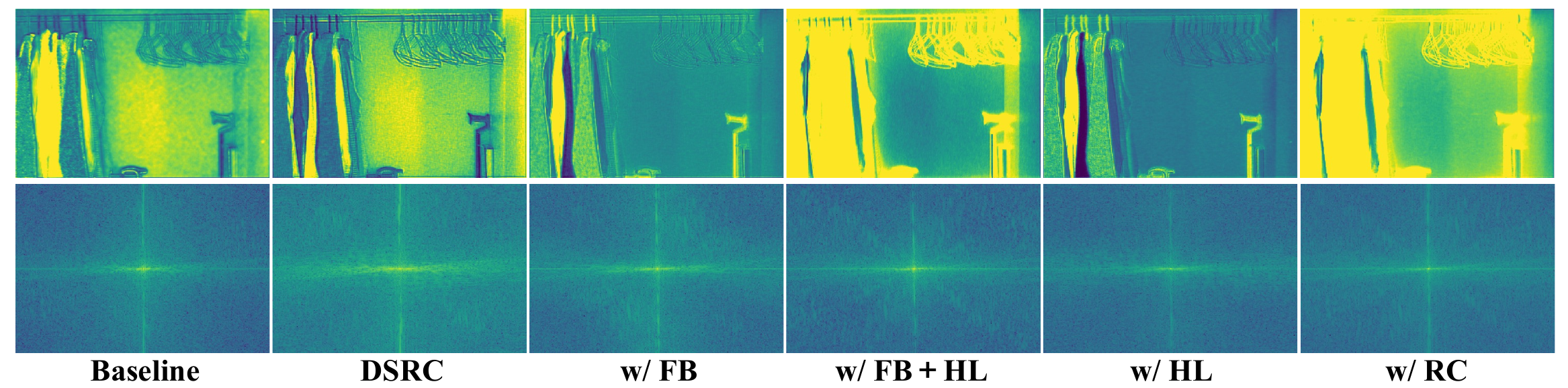}
    \caption{Feature-spectrum visualization of DSRC variants.}
    \label{fig:Feature Visualization}
    \end{figure}
    
    \subsection{Implementation Details}
We use three normalized radial bands ($B=3$) with boundaries $\{0,\tfrac18,\tfrac14,\tfrac12\}$. Under the normalized Euclidean radius in Eq.~(2), the partition separates low-, middle-, and high-frequency kernel components; in particular, it gives nonempty radial groups even for common $3\times3$ kernels. HL modulation uses $\sigma=0.8$ and the same spatial support as the corresponding convolution kernel.

Experiments cover LOL \cite{wei2018deep} for LLIE, UIEB \cite{li2019underwater} for UIE, and Set5 \cite{Bevilacqua2012Set5} for SR. We report PSNR, SSIM, LPIPS \cite{wang2004image,zhang2018unreasonable}, parameters, FLOPs, and latency. Methods marked with * replace compatible standard convolutions with DSRC, while all other training settings follow the official implementations of the evaluated backbones. All experiments are implemented in PyTorch and measured on an NVIDIA RTX 3090. Table~\ref{tab:comparison} covers four representative backbones per task, for a total of twelve models, allowing us to evaluate whether the same module transfers across distinct degradation types and network designs.

    \subsection{Results and Comparisons}

    \begin{table}[t]
    \centering
    \caption{Comparison of enhancement modules on LLIE using MobileIE as the baseline.}
    \label{tab:similar}
    \resizebox{\linewidth}{!}{
    \begin{tabular}{cccccccc}
        \toprule
        \textbf{Methods} & Venue & PSNR$\uparrow$ & SSIM$\uparrow$ & LPIPS$\downarrow$ & Params $\downarrow$  & FLOPs $\downarrow$  & Latency $\downarrow$  \\
        \midrule
        baseline \cite{yan2025mobileie}         & ICCV & 23.01 & 0.80 & 0.30 & 4.05   & 0.92   & 1.82 \\
        CSSC \cite{peng2025boosting}   & AAAI & 23.29 & 0.81 & 0.30 & 4.05   & 0.92   & 1.82 \\
        DyConv \cite{chen2020dynamic}  & CVPR & 23.31 & 0.81 & 0.30 & 49.99  & 14.14  & 26.98 \\
        FreqFusion \cite{chen2024frequency} & TPAMI & 23.72 & 0.80 & 0.29 & 19.36  & 3.27   & 5.39 \\
        \rowcolor{gray!15}
        \textbf{DSRC (Ours)} & -- & \textbf{23.74} & \textbf{0.82} & \textbf{0.29} & \textbf{4.05} & \textbf{0.92} & \textbf{1.82} \\
        \bottomrule
    \end{tabular}}
    \end{table}

        \begin{table}[t]
    \centering
    \caption{Ablation of frequency-band (FB), high/low-pass (HL), and residual compensation (RC) components.}
    \label{tab:ablation}
    \resizebox{0.98\linewidth}{!}{
    \begin{tabular}{ccc|ccc|ccc}
        \toprule
        \multicolumn{3}{c|}{Modulations} & \multicolumn{3}{c|}{MobileIE \cite{yan2025mobileie}} & \multicolumn{3}{c}{Complexity} \\
        \cmidrule(lr){1-3}\cmidrule(lr){4-6}\cmidrule(lr){7-9}
        FB & HL & RC & PSNR $\uparrow$ & SSIM $\uparrow$ & LPIPS $\downarrow$ & Params $\downarrow$ & FLOPs $\downarrow$ & Latency $\downarrow$ \\
        \midrule
        \checkmark &            &            & 23.40 (+0.39) & 0.81 (+0.01) & 0.29 (-0.01) & 4.05 (+0) & 0.92 (+0) & 1.82 (+0) \\
                   & \checkmark &            & 23.51 (+0.50) & 0.81 (+0.01) & 0.29 (-0.01) & 4.05 (+0) & 0.92 (+0) & 1.82 (+0) \\
                   &            & \checkmark & 23.33 (+0.32) & 0.81 (+0.01) & 0.29 (-0.01) & 4.05 (+0) & 0.92 (+0) & 1.82 (+0) \\
        \checkmark & \checkmark &            & 23.54 (+0.53) & 0.81 (+0.01) & 0.30 (-0.00) & 4.05 (+0) & 0.92 (+0) & 1.82 (+0) \\
        \rowcolor{gray!15}
        \checkmark & \checkmark & \checkmark & \textbf{23.74 (+0.73)} & \textbf{0.82 (+0.02)} & \textbf{0.29 (-0.01)} & \textbf{4.05 (+0)} & \textbf{0.92 (+0)} & \textbf{1.82 (+0)} \\
        \bottomrule
    \end{tabular}}
    \end{table}

    Table~\ref{tab:comparison} shows that DSRC improves PSNR across all twelve backbone-task configurations, with SSIM and LPIPS generally preserved or improved. After fusion, DSRC preserves the deployed convolution structure, parameter count, and theoretical FLOPs of the corresponding baseline, without introducing additional inference operators. Representative PSNR gains include +0.73 dB for MobileIE on LLIE, +0.28 dB for GHSUIR on UIE, and +0.16 dB for IDN on SR. The PSNR gains range from 0.23 to 0.89 dB on LOL, 0.23 to 0.67 dB on UIEB, and 0.03 to 0.17 dB on Set5.

    Table~\ref{tab:similar} compares DSRC with CSSC, DyConv, and FreqFusion under the same MobileIE framework. CSSC preserves baseline inference complexity but provides a smaller PSNR gain, whereas DyConv substantially increases parameters, FLOPs, and latency. FreqFusion approaches the PSNR of DSRC but introduces additional inference computation. DSRC attains the highest PSNR in this comparison while retaining baseline inference complexity, illustrating the intended balance between adaptability and deployment efficiency.

    Qualitative results in Fig.~\ref{fig:qualitative} are consistent with the quantitative trends. DSRC variants improve illumination and suppress noise in LLIE, recover more natural color and contrast in UIE, and preserve sharper structures in SR. These examples suggest that the learned spectral modulation adapts differently to task-specific degradation characteristics rather than applying a single fixed enhancement behavior.

    \subsection{Ablation Study}

Table~\ref{tab:ablation} evaluates the contributions of FB, HL, and RC on MobileIE. Each component independently improves PSNR, with HL achieving the largest individual gain (+0.50 dB). Combining FB and HL yields 23.54 dB, while adding RC further improves PSNR to 23.74 dB and SSIM to 0.82, with LPIPS reduced to 0.29. These results demonstrate the complementary effects of the three components, although their individual gains are not additive.

Fig.~\ref{fig:Feature Visualization} illustrates feature maps and their spectral representations for the baseline and DSRC variants. The displayed differences indicate that the components affect the learned representations in different ways. These visualizations complement the quantitative ablation. The full configuration yields a compact and structured response together with the best quantitative result.

\section{CONCLUSION}
\label{sec:CONCLUSION}
We presented DSRC, a re-parameterizable convolution that combines kernel-level frequency bands, high/low-pass modulation, and residual compensation. Unlike feature-level frequency modules, DSRC performs spectral shaping on convolution kernels and analytically fuses all training-time components into a standard kernel. Experiments across LLIE, UIE, and SR show consistent PSNR improvements over twelve backbone–task configurations while preserving baseline deployed parameter counts and theoretical FLOPs.

\clearpage

\footnotesize
\bibliographystyle{IEEEbib}
\bibliography{strings,refs}

\end{document}